\documentclass{article}
\usepackage{spconf,amsmath,graphicx}
\usepackage{multirow}
\usepackage{amsfonts}
\usepackage{booktabs}

\title{Self-supervised learning with Cross-Modal transformers for emotion recognition}
\name{Aparna Khare, Srinivas Parthasarathy, Shiva Sundaram}
\address{Amazon.com, Sunnyvale, CA}
\begin{document}
%
\maketitle
\begin{abstract}
 Emotion recognition is a challenging task due to limited availability of in-the-wild labeled datasets.\ Self-supervised learning has shown improvements on tasks with limited labeled datasets in domains like speech and natural language.\ Models such as BERT learn to incorporate context in word embeddings, which translates to improved performance in downstream tasks like question answering.\ In this work, we extend self-supervised training to multi-modal applications.\ We learn multi-modal representations using a transformer trained on the masked language modeling task with audio, visual and text features.\ This model is fine-tuned on the downstream task of emotion recognition.\ Our results on the CMU-MOSEI dataset show that this pre-training technique can improve the emotion recognition performance by up to 3\% compared to the baseline. 
\end{abstract}
\section{Introduction}
\label{sec:intro}

Human communication is inherently multi-modal in nature. Our expressions and tone of voice augment verbal communication.\ This can include vocal features like speaking rate, intonation and visual features like facial expressions \cite{burgoon2016nonverbal}. Non-verbal communication is important for tasks that involve higher level cognitive expressions like emotions \cite{schuller2011avec}, persuasiveness \cite{nojavanasghari2016deep} and mental health analysis \cite{haque2018measuring}. We focus on a multi-modal approach to emotion recognition because humans fundamentally express emotions verbally using spoken words \cite{chung2007psychological}, as well as with acoustic signals \cite{banse1996acoustic} and visual expressions \cite{cohn2006foundations}.

Getting large-scale labeled datasets for emotion recognition can be challenging.\ Our primary motivation for this paper is to study effective utilization of large unlabeled datasets to improve performance of multi-modal emotion recognition systems.\ The signals we consider are speech, visual information and spoken text.\ Our motivation stems from the popular use of pre-trained models in natural language, speech and visual understanding tasks to circumvent data limitations.\ BERT is a popular model for natural language understanding \cite{devlin2018bert} that was trained using self-supervision.\ Devlin et al. use the masked language modeling (LM) task on the Wikipedia corpus for pre-training.\ The model was successfully fine-tuned to improve performance on several tasks like question answering and the general language understanding evaluation benchmarks \cite{devlin2018bert}. Self-supervised learning has also been successfully applied to speech based applications.\ Schneider et al.\ in \cite{schneider2019wav2vec} use unsupervised pre-training on speech data by distinguishing an audio sample in the future from noise samples.\ Fine-tuning this model shows state of the art results on automatic speech recognition (ASR). Liu et al.\ show in \cite{liu2020mockingjay} that a BERT-like pre-training approach can be applied to speech.\ By predicting masked frames instead of masked words, the performance on tasks like speaker recognition,\ sentiment recognition and phoneme classification can be improved. For emotion recognition, Tseng et al.\ show in \cite{tseng2019multimodal} that text-based self-supervised training can outperform state of the art models. The authors use a language modeling task, that involves predicting a word given its context, to pre-train the model.\ Another area of work that has leveraged unlabeled data is detection and localization of visual objects and spoken words in multi-modal input.\ Harwath et al.\ in \cite{harwath2017learning,harwath2020jointly} train an audio-visual model on an image-audio retrieval task.\ The models are trained to learn a joint audio-visual representation in a shared embedding space.\ This model can learn to recognize word categories by sounds without explicit labels.\ Motivated by the success of these approaches, we study if similar methods can be applied to multi-modal emotion recognition.\ To the best of our knowledge, a joint self-supervised training approach using text, audio and visual inputs has not been well explored for emotion recognition. 

Multi-modal emotion recognition models have been well studied in literature and typically outperform uni-modal systems \cite{liang2018computational}.\ These models need to combine inputs with varying sequence lengths.\ In video, the sequence lengths for audio and visual frames differ from the length of text tokens by orders of magnitude.\ There has been considerable prior work in fusing multi-modal features. Liang et al.\ in \cite{liang2018computational} studied multiple fusion techniques for multi-modal emotion recognition and sentiment analysis.\ Their methods included early and late fusion of modalities, and a dynamic fusion graph based network.\ They showed that the graph fusion model outperforms other methods.\ Early fusion and graph fusion techniques both require alignment between various modalities.\ Late fusion can be performed without alignment, but does not allow interaction of features from different modalities at the frame level.\ To overcome this limitation,\ Tsai et al.\ introduce the cross-modal transformer in \cite{tsai2019multimodal}.\ It scales the features using cross-modal attention.\ In the process, the modalities are projected into sequences of equal lengths, eliminating the need for any alignment.\ This architecture has been successfully applied to problems like emotion recognition, sentiment analysis \cite{tsai2019multimodal,khare2020multi} and speech recognition \cite{paraskevopoulos2020multiresolution}.\ Recently, another transformer-based method to combine multi-modal inputs was introduced by Rahman et al. in \cite{rahman2020integrating}, which uses a multi-modal adaptation gate.

In this paper, we propose using the same pre-training scheme as BERT, but extend it to a model that uses audio, visual and text inputs. We discuss the relevance of this approach in Section \ref{subsec:sst}.\ The multi-modal representations learned in pre-training are fine-tuned for emotion recognition.\ We evaluate the efficacy of the pre-training approach.\ We also perform experiments to understand the importance of each modality on the CMU-MOSEI dataset and provide case-studies to interpret the results. 

This paper is organized as follows.\ In Section \ref{sec:selfsupervised} we describe our model architecture and the self-supervised approach for pre-training, along with further motivation for the self-supervised learning we choose.\ In Section \ref{experiments}, we discuss the training setup and data.\ We present our results and analysis in Section \ref{results} and conclude in Section \ref{discussion}.

\section{Self-supervised training with cross-modal transformers}
\label{sec:selfsupervised}

\subsection{Model architecture}
\label{subsec:modelarc} 
Not all information in a given sequence is equally important for emotion recognition.\ If we consider visual inputs, emotionally relevant cues may appear only in certain frames. Similarly, each spoken word in the sentence does not contribute equally to the expressed emotion. Given this nature of the sequence recognition problem, transformer-based models are a good choice for extracting a fixed length representation for emotion recognition. 

We use the cross-modal transformer for emotion recognition since it showed state of the art results on sentiment analysis \cite{tsai2019multimodal}.\ We chose a modified version of the proposed model and will describe it in this section.\ The architecture allows each sample from each modality to interact with each sample from each other modality, providing the benefits of low-level fusion.\ It also projects all the sequences into equal lengths which allows for frame level late fusion after the transformation.

\begin{figure*}[t]			
\centering
\includegraphics[width=\textwidth]{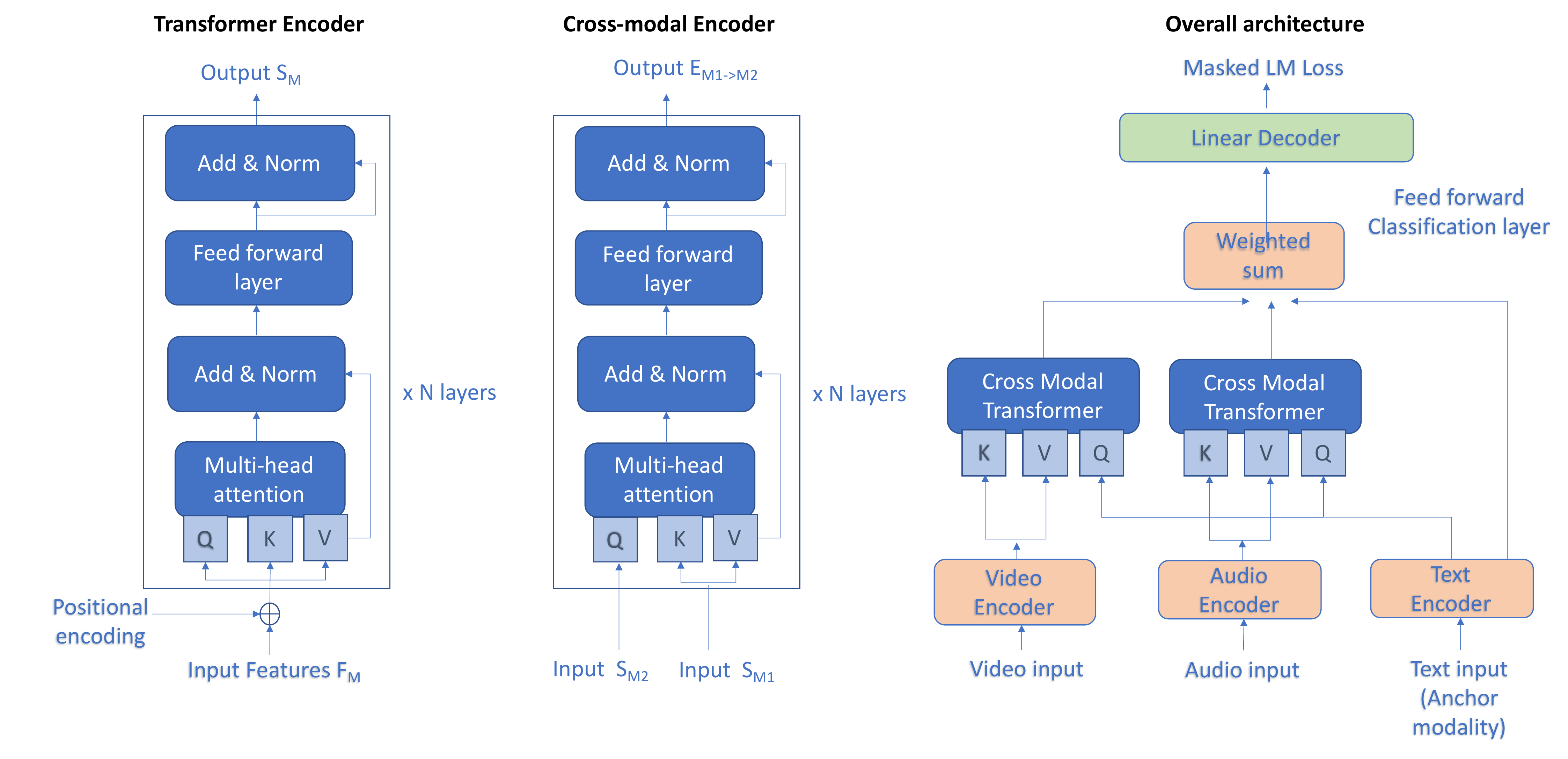}
 \caption{Cross-modal transformer based self-supervised learning architecture. The figure shows the self-attention based transformer encoder layers, the cross-modal attention encoder module and the overall architecture of our self-supervised model.}
 \label{fig:multitask}
\end{figure*}
Our overall architecture is shown in Figure \ref{fig:multitask}.\ The transformer model trained for emotion recognition allows for attending to specific input features (visual frames, words, speech segments) that are relevant to the task \cite{tsai2019multimodal}. The first part of our model architecture achieves this by using self-attention based transformer encoder for individual modalities.\ We add positional embeddings to the input features as discussed in \cite{vaswani2017attention}. Intuitively, positional embeddings would be useful in the task because for extracting the context from the input, the order of the words matter.\ We did not study the importance of positional embeddings since our work is focused on self-supervised learning.\ These features are processed by the transformer encoder.\ The architecture of the encoder layers is identical to \cite{vaswani2017attention} and is shown in Figure \ref{fig:multitask}.\ The transformer encoder consists of N layers.\ The first operation in each layer transforms the input into keys, queries and values.\ If the input to a given layer for modality $\mathbb{M}$ is represented by $F_{\mathbb{M}}$, then the query $Q_{\mathbb{M}}$, the key $K_{\mathbb{M}}$ and the value $V_{\mathbb{M}}$ for the corresponding modality is computed as 

\begin{equation}
\begin{array}{l}
Q_{\mathbb{M}} = W_{q{\mathbb{M}} }(\mathbb{F}_{{\mathbb{M}} } )\\
K_{\mathbb{M}}  = W_{k{\mathbb{M}} }(\mathbb{F}_{\mathbb{M}} )\\
V_{\mathbb{M}}  = W_{v{\mathbb{M}} }(\mathbb{F}_{\mathbb{M}} )\\
\end{array}
\end{equation}
where $W_{\alpha\beta}$ represents a linear projection.\ After obtaining the keys, queries and values, the self-attention layer scales the value $V_{\mathbb{M}}$. The output of the attention layer, represented by $A_{\mathbb{M}}$, is computed as
\begin{equation}
A_{\mathbb{M}} = \text{softmax}(\frac{Q_{\mathbb{M}}K^T_{\mathbb{M}}}{\sqrt{d}})V_{\mathbb{M}}
\end{equation}
where $d$ denotes the dimensionality of the keys.\ In practice, we use the multi-head version of the scaled dot-product attention that uses $k$ scaled dot-product attention heads.\ The final output of the transformer encoder layer, $S_{\mathbb{M}}$, is computed as following

\begin{equation}
\begin{array}{l}
O_{\mathbb{M}} = LayerNorm(V_{\mathbb{M}} + A_{\mathbb{M}})\\
S_{\mathbb{M}} = LayerNorm(O_{\mathbb{M}} + FeedForward(O_{\mathbb{M}}))
\end{array}
\end{equation}
where $O_{\mathbb{M}}$ is the normalized output after adding a residual connection from $V_{\mathbb{M}}$ to the output of the scaled dot-product attention layer. We use $N$ encoder layers to obtain the self-attended outputs $S_{\mathbb{A}}$, $S_{\mathbb{V}}$ and $S_{\mathbb{T}}$ from the audio, visual and text modalities respectively.

Next, we combine the uni-modal transformer encoder outputs, $S_{\mathbb{A}}$, $S_{\mathbb{V}}$ and $S_{\mathbb{T}}$, to learn the final multi-modal representation for emotion recognition.\ This is done by the cross-modal transformer, which computes the attention map between features from two different modalities $\mathbb{M}_1$ and $\mathbb{M}_2$. The cross-modal attention allows for increasing attention weights on features that are deemed important for emotion recognition by more than one modality. This property was shown in \cite{tsai2019multimodal}.

The output $E_{\mathbb{M}_1 -> \mathbb{M}_2}$ from the cross-modal transformer is computed as:

\begin{equation}
E_{\mathbb{M}_1 -> \mathbb{M}_2} = CM(Q_{E_{\mathbb{M}_2}},K_{E_{\mathbb{M}_1}},V_{E_{\mathbb{M}_1}})
\end{equation}
where $CM$ denotes the transformations applied by the cross-modal transformer, shown in Figure \ref{fig:multitask}. The cross-modal transformer is identical to the transformer encoder, with one exception.\ The key and value matrices ($K_{E_{\mathbb{M}_1}}$\ and $V_{E_{\mathbb{M}_1}}$) are obtained from the encoded output $S_{\mathbb{M}_1}$.\ The query $Q_{E_{\mathbb{M}_2}}$, which provides contextualization, is obtained from the encoded output of modality $S_{\mathbb{M}_2}$.\ The keys, queries and values are obtained from the outputs of the corresponding uni-modal transformers by using fully connected layers, similar to the uni-modal transformer encoders described above.\ The final encoder output $E_{\mathbb{A}\mathbb{V}\mathbb{T}}$ is the weighted sum of the encoded output of the text modality and the cross-modal transformer output attending to the audio and visual inputs using the context query from the text domain.
\begin{equation}
E_{\mathbb{A}\mathbb{V}\mathbb{T}} = w_1 \cdot E_\mathbb{T} + w_ \cdot E_{\mathbb{A} -> \mathbb{T}} + w_3 \cdot E_{\mathbb{V} -> \mathbb{T}}
\end{equation}
For our experiments, we used a fixed weight of 0.33 for $w_1$, $w_2$ and $w_3$.\ The classification consists of average pooling followed by a linear layer that maps the representation into emotion classes. 

Our model differs from the architecture in \cite{tsai2019multimodal} in two ways. Instead of using convolutional layers to increase temporal awareness, we use uni-modal encoders to encode the various modalities.\ We used a second modification by limiting the cross-modal transformer layers to only compute attention between audio and text, and visual input and text as described above.\ We call text the anchor modality in this architecture. For this study, we chose text as the anchor modality for ease of the self-supervised task. This allows us to train the self-supervised model without a decoder. For comparison, we tried training our baseline models with both audio and video as anchors and the results were similar. Note that our focus in this work is to study the impact of self-supervised training, and not to tune the model architecture for emotion recognition. Hence, we chose the simplified model architecture for all the experiments. Our work does not compare our model to the one proposed in \cite{tsai2019multimodal} and we will do so in our future work.	

\subsection{Self-supervised training}
\label{subsec:sst}
Our primary motivation for this study is to understand how to leverage large unlabeled multi-modal datasets to learn representations which can be fine-tuned for emotion recognition.\ We consider what self-supervised task would be relevant given our downstream task of interest.\ We note that spoken words are one of the strongest expressions of human emotion.\ This has been studied in psychological literature \cite{chung2007psychological}, and also in the emotion recognition literature.\ Embeddings like ELMo, that encode language context, can be applied successfully to the emotion recognition task \cite{tseng2019multimodal}. Such representations from text have been learned successfully with self-supervised tasks like skip-gram, continuous bag-of-words model \cite{mikolov2013efficient} and more recently using the masked LM task \cite{devlin2018bert}.\ We extend this work by learning representations that encode context using the text input, as well as the audio and visual inputs.

We choose the masked LM task to train the model, similar to the BERT model \cite{devlin2018bert}. We propose to predict words by looking at audio and visual context in addition to the context words around the masked input.\ Intuitively, the auxiliary information present in visual expressions and audio features like intonation would provide relevant input for predicting the masked words.\ For example, consider the phrase ``This movie is [MASK]'' as an input to the model. The [MASK] word could be predicted as ``amazing'' if the audio and visual features show that speaker is happy.\ Alternatively, the prediction could be ``terrible'' if the speaker seems discontent while talking. This information cannot be derived from text only input.\ We posit that the latent representations learned using the masked LM task with multi-modal input will not just encode context, but also the relevant emotional information which could be used to predict those words.	

For training, we mask 15\% of the input words and the audio and visual features corresponding to those words.\ The word boundaries are obtained using an existing ASR system. More details on the ASR system used will be discussed in Section \ref{datatraining}.\ The model predicts the words at each sequence, and the loss is computed only for the words that were masked in the input sequence.\ Instead of providing a mask token as input for masked words, we choose to set the masked input for all modalities to zero.\ We are able to do so as we use GLoVe embeddings to represent the text input instead of learning an embedding layer.\ Similarly, to mask the audio and visual inputs for the corresponding masked words, we replace the input features with zeros.\ For this task, we replace the average pooling and linear classifier described in Section \ref{subsec:modelarc} with a linear layer of output size equal to the model vocabulary. Since the encoder layer uses bi-directional attention, we do not need a decoder to attend to past predictions from the model. In addition, since the encoder output length is equal to the sequence length of the input text, we do not require a transformer decoder layer. This allows training to be simplified and was one of the reasons we chose the model architecture.

For the loss function, we use a full softmax loss as well as noise contrastive estimation (NCE) \cite{gutmann2012noise} to train our models.\ NCE has been used successfully to learn the inverse language modeling task, that involves predicting the context words given a word.\ Minh et al.\ show in \cite{mnih2013learning} that NCE can reduce computation by estimating the normalization factor for computing softmax using noise samples. For a task which has a similar vocabulary size as our dataset, they demonstrated a reduction of up to 50\% in training time. We compare the models trained with NCE loss with the full softmax loss. This would inform if the multi-modal transformer can be trained with similar accuracy but more efficiently. Our implementation is exactly the same as \cite{mnih2013learning}, except we use a normalization factor of $\text{vocabulary size}$ which we found to be critical for training our model.

\section{Experimental setup}
\label{experiments}

\subsection{Dataset and training details}
\label{datatraining}
Our setup involves first training the cross-modal transformer on the masked LM task on a large dataset, followed by fine-tuning for emotion recognition. For pre-training, we utilize the publicly available VoxCeleb2 dataset \cite{Chung18b}.\ We chose this dataset since it provides all the modalities we are interested in and is sufficiently large (1.1 million videos in the train partition).\ More importantly, this data is emotion rich, as shown in \cite{Albanie18a}.\ This dataset does not provide text transcriptions.\ We used a TDNN ASR model trained with the standard Kaldi recipe on the Librispeech dataset to get transcriptions \cite{peddinti2015time}.\ We use 40-dimensional Log-Filter bank energy (LFBE) features using a 10ms frame rate to represent the audio input. Visual frames are represented by 4096-dimensional features extracted from the\ VGG-16 model.\ 300-dimensional GloVe embeddings represent the text input.\ We chose to use GLoVe embeddings for this task instead of an embedding layer because our dataset has a limited number of sentences and vocabulary. The vocabulary size for this dataset as obtained from ASR transcriptions is 88000. Using GloVe embeddings allows the model to take advantage of pre-trained embeddings trained on billions of words \cite{pennington2014glove}.\ The disadvantage is the inability to handle out of vocabulary words, which we ignore for all our experiments. 

The model is pre-trained using pytorch with the learning schedule described in \cite{dong2018speech}.\ We stack 5-frames of the LFBE features for a final audio feature dimensionality of 200. This was done to reduce the memory requirements for training the model. We select only the English language videos from VoxCeleb2 for training. The filtering is done by selecting a heuristic threshold on the likelihood scores from the ASR decoder. For all our experiments, we use only the dev portion of the VoxCeleb2 dataset. Our final training dataset consists of 978k utterances from 4820 speakers.\ We use the architecture described in Section \ref{subsec:modelarc} to train the model. The model has keys, values and queries of dimension 512, 4 encoder layers, feed forward layer of dimension 200 and 4 attention heads for both the uni-modal and cross-modal encoders. We trained the model for 20 epochs and chose the final model with the lowest loss on a held-out set.

For evaluating performance on the emotion recognition task, we fine-tune the model on the CMU-MOSEI dataset \cite{liang2018computational}.\ It is the largest publicly available multi-modal dataset for emotion recognition with natural conversations.\ It contains 23,453 single-speaker video segments from YouTube.\ The clips have been manually transcribed and annotated for sentiment and emotion.\ The dataset consists of 6 emotions; happy (12135 examples), sad (5856 examples), angry (4903 examples), disgust (4208 examples), surprise (2262 examples) and fear (1850 examples).\ The labels for each class are on a Likert scale of $[0,3]$. We convert the labels into binary targets. A clip is assigned a 0 label if the score on the Likert scale is 0, and 1 otherwise.\ A greater than 0 score on the Likert scale represents the presence of the specific emotion, and 0 the absence of the emotion.\ This was reflected in the binary interpretation that we chose. For fine-tuning the model, we remove the decoder layer for the masked LM task and add the average pooling and decoder layer for emotion recognition.\ Each example in this dataset can be labeled with the presence of multiple emotions. Therefore, we use a sigmoid output for each of the 6 nodes in the output layer to get the probability for each emotion.\ The positive and negative examples for various emotions in the dataset are imbalanced.\ During training, we weigh the loss for the each training sample appropriately to ensure that the positive and negative examples across all emotions contribute equally to the loss.

\begin{table*}[tp]
\begin{center}
\caption{Emotion recognition results on the CMU-MOSEI task. The 95\% confidence interval for all metrics is less than $\pm 1.4$}
\label{tab:moseibaseline}
\resizebox{\textwidth}{!}{\begin{tabular}{c c c c c c c c c c c c c c c}
\toprule
\multirow{2}{*}{\parbox{0\linewidth}}{\bf Model} &\multicolumn{2}{c}{\bf Happy} &\multicolumn{2}{c}{\bf Sad} &\multicolumn{2}{c}{\bf Anger} &\multicolumn{2}{c}{\bf{Surprise}} &\multicolumn{2}{c}{\bf{Disgust}} &\multicolumn{2}{c}{\bf {Fear}}&\multicolumn{2}{c}{\bf {Average}}\\ 
\multirow{2}{*}{}& \multicolumn{6}{c}{}\\
 & WA & F1 & WA  & F1 & WA & F1& WA & F1 & WA & F1 & WA & F1 & WA & F1 \\ 
\midrule
M-ELMo + NN \cite{tseng2019multimodal}(A+T) & 67.0	&65.2	&63.1	&72.0	&65.8	&\bf{74.7}&	\bf{63.8}	&83.3	&74.2&81.7	& \bf{63.2} &85.1	&66.2	&77.0 \\

Graph-MFN \cite{liang2018computational} & 66.3 & 66.3 & 60.4 & 66.9 &  62.6 & 72.8 & 53.7 & 85.5 &  69.1 & 76.6 & 62.0  & \bf{89.9} &62.3 &76.3\\
\\
Transformer (baseline)  & \bf{67.4} & \bf{67.1} &\bf{64.6}& \bf{72.5} &\bf{68.2}& \bf{74.7} &  62.9 &\bf{88.1} & \bf{74.8}& \bf{82.4} &61.5& 86.5&\bf{66.6} & \bf{78.5} \\
\midrule
\midrule
Transformer with pre-training and full softmax loss & \bf{68.1} & 68.1  & \bf{65.1} & 72.1	& 67.0  &  74.4&	\bf{65.1} & 88.0 & 74.5 & 82.3	& \bf{64.5} & 86.4&	\bf{67.4}	&78.6 \\

Transformer with pre-training and NCE loss &  \bf{68.1} &  \bf{68.2}& 	64.3&  72.4	& 67.3 &  \bf{74.8	}&   \bf{65.1}&  87.7& 	73.6&  82.4& 	63.0 &  86.6	&  66.9	&  \bf{78.7}\\
\bottomrule
\end{tabular}}
\end{center}
\end{table*}

\begin{table}
\begin{center}
\caption{Ablation studies with the baseline model. Note that the text modality cannot be ablated with this architecture.}
\vspace{4mm}
\label{tab:ablation}
\begin{tabular}{c c c}
\toprule
& \bf{WA} & \bf{F1}\\
 \midrule
 Text only & 64.5 & 76.7  \\

Audio + Text &  65.3 & 78.1 \\

Video + Text  & 65.3 & 78.5 \\

Audio + Video + Text  & 66.6 & 78.5\\
\bottomrule
\end{tabular}
\end{center}
\vspace{-2mm}
\end{table}
\section{Results}
\label{results}
We use the weighted accuracy (WA) and F1-score for each emotion as the metrics for the task.\ We also report average of these two metrics over the 6 emotions, keeping in line with prior work \cite{liang2018computational}. For evaluating the baseline model, we follow the procedure in \cite{tseng2019multimodal}. The model is randomly initialized and trained 10 different times. The best model is chosen based on the average of the weighted accuracy and F1-scores over all the emotions on the dev set over the 10 runs. 

\subsection{Results on emotion recognition}

Table \ref{tab:moseibaseline} shows the results of our experiments and state of the art results on the same dataset from other publications.\ The transformer baseline outperforms or is comparable to published results for most of the metrics.\ Our model shows a 2.4\% absolute improvement in the weighted accuracy of the anger emotion and a 2.6\% absolute improvement in the F1-score of the surprise emotion. We observe a degradation in the weighted accuracy of the fear emotion.\ This comparison with other state of the art models is pertinent for the rest of our work as we want to build upon a strong baseline model.

The next set of results in Table\ \ref{tab:moseibaseline} are with the pre-trained model on the VoxCeleb2 dataset, fine-tuned for emotion recognition.\ Our results show up to 3\% absolute improvement in the weighted accuracy of 4 out of the 6 emotions, with a slight degradation in the weighted accuracy of the anger emotion.\ The average weighted accuracy over all emotions improves by 0.8\%. The weighted accuracy of the surprise and fear emotions improve by 2.2\% and 3\% absolute respectively. The 95\% confidence intervals of these emotions don't overlap with the baseline, demonstrating the statistical significance of the results. The F1-score is comparable to the baseline for all emotions other than happy, where we see a 1\% absolute improvement.\ The model trained using the NCE optimization has similar improvements. It shows that we can achieve the same improvements with a lower computational cost of training.\ These results validate our hypothesis that we can effectively leverage a large unlabeled multi-modal dataset to improve results on emotion recognition using self-supervised pre-training. 

In order to understand the impact of ASR errors on the model,\ we generated transcriptions on the CMU-MOSEI dataset using a commercial ASR system.\ The word error rate of the machine-generated transcriptions was 29\%. We then re-evaluated the performance of our baseline model with ASR based transcriptions instead of the transcriptions provided as part of the dataset. We did not observe a degradation in emotion recognition performance. Note that for this experiment, the baseline model was trained with the original transcriptions. ASR errors have been studied well in literature and it has been shown that the top contributors to errors are shorter words like 'on', 'was', 'in' etc. \cite{stolcke2017comparing}. These words do not contribute to emotion expression, which would explain the observations we made. 

\subsection{Analysis and case studies}
We analyze the results to understand the contribution of each modality towards accuracy.\ We look at predictions from the baseline model with missing inputs from select modalities. Note that we cannot ablate the text input. The output of the cross-modal transformers will be 0 if the text input is 0 since the attention maps will be all zeros.\ For subjective analysis with missing text input, we trained a baseline model with audio as the anchor modality.\ As noted in Section \ref{subsec:modelarc}, the choice of anchor modality does not change the performance of the baseline model.\ We describe our subjective analysis below.

The first example we observe is ID ``HeZS2-Prhc[8]'' in the dataset.\ From visual inspection, the video shows that the speaker is laughing, which conveys a happy emotion. However, the speaker is talking about the cost for drugs and its impact on communities. This is why the visual modality is the key to accurately predicting the emotion, and the model is not able to classify the emotion as happy with text input alone.\ On the contrary, the second example, ID ``10219[11]'', shows the speaker with a neutral face and a neutral tone of voice. The speaker is talking about a positive movie review, which leads to the text classifying the emotion as happy. The model was not able to classify the emotion in this example as happy without the text input. In the third example, ID ``-9y-fZ3swSY[1]'', the speaker is talking about a neutral topic with a slightly positive face, but in a very positive tone of voice.\ The model predicts that the speaker in this video is happy only when the audio features are present.\ This subjective analysis shows the importance of multi-modal features in human communication, and how each of them contribute to emotion recognition. 

Next,\ we show the overall results with each missing modality in Table \ref{tab:ablation}.\ Adding audio and visual input along with text improves both metrics by 2\% absolute.\ The results show that with text alone, we can recover most of the baseline performance. The subjective examples, however, suggest that for several cases, other modalities are required for accurate prediction.\ Therefore, the importance of text should not be generalized for the problem of emotion recognition in-the-wild.\ However, for the CMU-MOSEI dataset, text is the most important modality for emotion recognition.\ To analyze this, we look at the distribution of topics in the dataset. The 5 most frequent topics are: reviews (16.2\%), debate (2\%), consulting (1.8\%), financial (1.8\%) and speech (1.6\%). For these topics, the perceived emotion by a human annotator is strongly based on what is being said. This would explain why text is the most important input. We posit that for more diverse topics, specifically involving human to human communication, the other modalities would start to gain importance for recognizing the emotions accurately.

\section{Conclusion}
\label{discussion}
In this paper, we present state of the art results on the emotion recognition task using the cross-modal transformer on the CMU-MOSEI dataset.\ We utilize a BERT-like pre-training scheme using audio, visual and text inputs.\ We use the VoxCeleb2 dataset to pre-train the model and fine-tune it for the emotion recognition task.\ We demonstrate up to a 3\% improvement over the baseline with the fine-tuned model. We presented our subjective analysis on the contribution of various modalities to emotion recognition.\ We also show results with missing input modalities to understand the importance of each modality for the emotion recognition task.

For our future work, we propose to initialize the text encoder with a text-only model like BERT, before multi-modal self-supervised training.\ VoxCeleb2 dataset, although large in terms of number of hours of video, is smaller when compared to the Wikipedia corpus which has billions of words. Taking advantage of a larger text-only corpus could provide improvements.\ We would also like to experiment with adapting the model on the CMU-MOSEI dataset.\ Both the VoxCeleb2 and CMU-MOSEI datasets are obtained from YouTube, but there could be domain mismatch between the two datasets. Adapting could help bridge the mismatch.\ We would also like to explore weak labels to adapt the pre-trained representations for the downstream task.\ Tseng et al.\ showed in \cite{tseng2019unsupervised} that weakly supervised labels can be used to effectively bias the embeddings learned by a pre-trained model. Even though we study the impact of ASR errors on emotion recognition, we do not know how these errors impact the self-supervised training. We would like to study that in the future. As noted before, our model architecture doesn't allow ablation of text. For our future work, we will focus on overcoming that limitation.

\bibliographystyle{IEEEtran}

\bibliography{mybib}

\end{document}